%% file: acl2021.tex
\documentclass[11pt,a4paper]{article}
\usepackage[hyperref]{acl2021}
\usepackage{times}
\usepackage{latexsym}
\usepackage{graphicx}
\usepackage{bm}
\usepackage{multirow} 
\usepackage{amsmath}
\usepackage{amssymb}
\usepackage{amsfonts}
\usepackage{booktabs}

\usepackage{microtype}

\aclfinalcopy % Uncomment this line for the final submission
\def\aclpaperid{445} %  Enter the acl Paper ID here

\title{SIRE: Separate Intra- and Inter-sentential Reasoning for\\Document-level Relation Extraction}

\author{
  Shuang Zeng$^{1,3}$,
  Yuting Wu$^{1,2}$\and
  Baobao Chang$^{1}$\thanks{\;\;Corresponding author.} \\
  $^{1}$The MOE Key Laboratory of Computational Linguistics, Peking University, China \\
  $^{2}$Wangxuan Institute of Computer Technology, Peking University, China\\
  $^{3}$School of Software and Microelectronics, Peking University, China\\
  \texttt{
    \{zengs,wyting,chbb\}@pku.edu.cn
  }
}
\date{}

\begin{document}
\maketitle

% \renewcommand{\thefootnote}{\fnsymbol{footnote}} %将脚注符号设置为fnsymbol类型，即特殊符号表示
% % \footnotetext[1]{Equal contribution.} %对应脚注[1]
% % \footnotetext[2]{Corresponding author.} %对应脚注[2]
% \footnotetext[1]{Corresponding author.} %对应脚注[1]
% % \renewcommand{\thefootnote}{\fnsymbol{footnote}}

\input{files/abstract.tex}
\input{files/introduction.tex}

\input{files/model.tex}

\input{files/experiments.tex}

\input{files/related_work.tex}
\input{files/conclusion.tex}

\section*{Acknowledgments}
The authors would like to thank the reviewers for their thoughtful and constructive comments.
This paper is supported by the National Key R\&D Program of China under Grand No.2018AAA0102003, the National Science Foundation of China under Grant No.61936012 and 61876004.

\bibliographystyle{acl_natbib}
\bibliography{anthology,acl2021}

\input{files/appendix.tex}

\end{document}

%% file: files/abstract.tex
\begin{abstract}
Document-level relation extraction has attracted much attention in recent years. It is usually formulated as a classification problem that predicts relations for all entity pairs in the document. However, previous works indiscriminately represent intra- and inter-sentential relations in the same way, confounding the different patterns for predicting them. Besides, they create a document graph and use paths between entities on the graph as clues for logical reasoning. However, not all entity pairs can be connected with a path and have the correct logical reasoning paths in their graph. Thus many cases of logical reasoning cannot be covered. This paper proposes an effective architecture, SIRE, to represent intra- and inter-sentential relations in different ways. We design a new and straightforward form of logical reasoning module that can cover more logical reasoning chains. Experiments on the public datasets show SIRE outperforms the previous state-of-the-art methods. 
% \footnote{We will release our code upon acceptance.}
Further analysis shows that our predictions are reliable and explainable. Our code is available at \url{https://github.com/DreamInvoker/SIRE}.

\end{abstract}
% Besides, they still use an implicit way to tackle logical reasoning among inter-sentential relations and omit the nature of the logical reasoning process. 

%% file: files/introduction.tex
\section{Introduction\label{sec:introduction}}

% 引入篇章级关系抽取任务。
Relation Extraction (RE) is an important way of obtaining knowledge facts from natural language text. Many recent advancements~\citep{inter, EoG, DocRED-paper, LSR, GAIN, GLRE} manage to tackle the document-level relation extraction (doc-level RE) that extracts semantic relations among entities across multiple sentences. Due to its strong correlation with real-world scenarios, doc-level RE has attracted much attention in the field of information extraction. 

\begin{figure}
    \centering
    \includegraphics[width=1.0\linewidth]{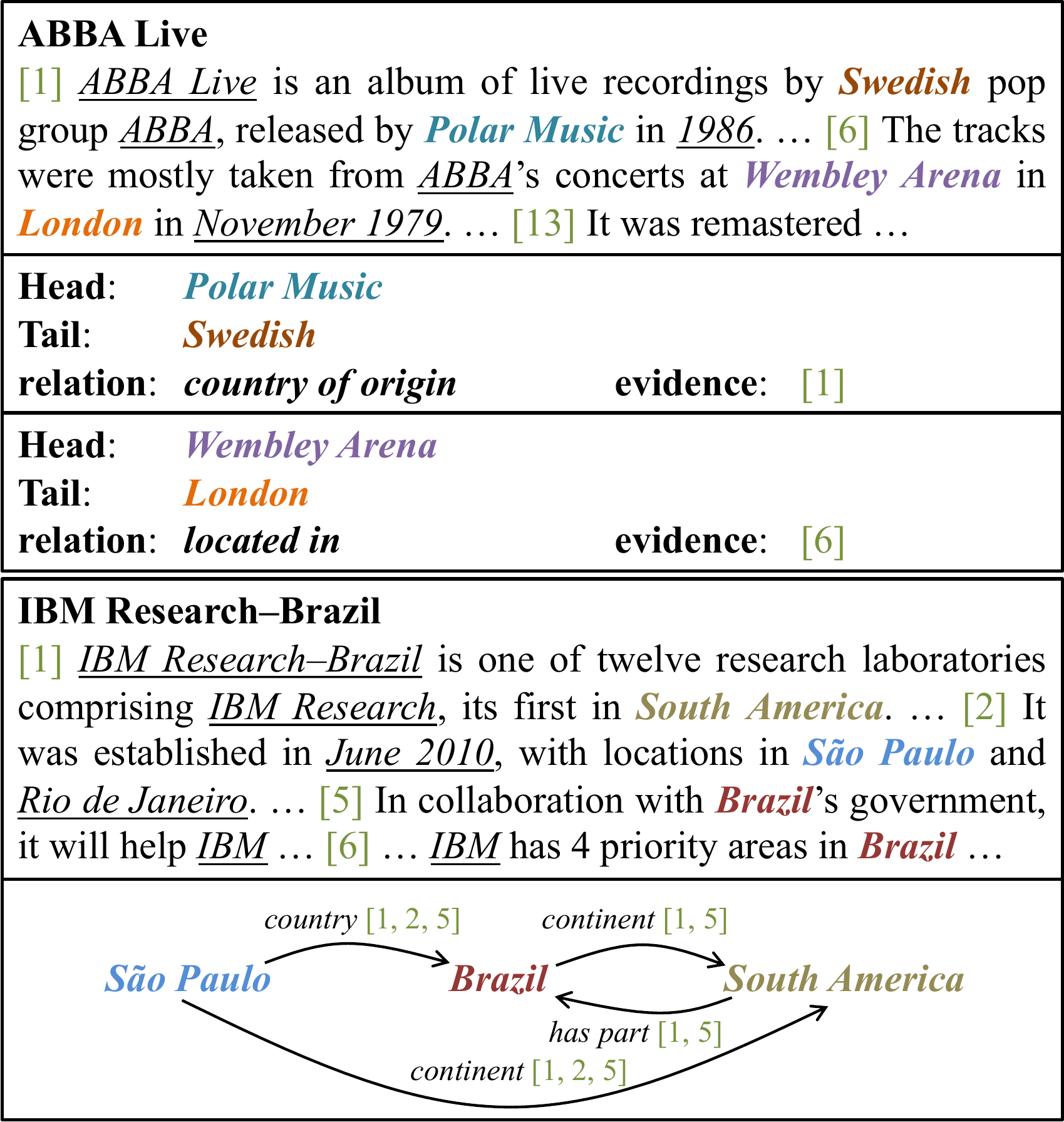}
    \caption{Two examples from DocRED~\citep{DocRED-paper} for illustration of intra- and inter-sentential relations. Sentence numbers, entity mentions, and supporting evidence involved in these relation instances are colored. Other mentions are underlined for clarity.}
    \label{fig:running-example}
\end{figure}

% 引出句内和句间关系两个概念。
% Previous works view this task as a classification problem that predicts possible relations for all entity pairs, using the information from the entire document. 
% Moreover, they use the performance on two different types of relations to evaluate a model: intra-sentential relation and inter-sentential relation, as illustrated in Figure~\ref{fig:running-example}. When two entities have mentions co-occurred in the same sentence, they may express intra-sentential relations. Otherwise, they may express inter-sentential relations.
The doc-level RE task is usually formulated as a classification problem that predicts possible relations for all entity pairs, using the information from the entire document.
It has two different kinds of relations: intra-sentential relation and inter-sentential relation. We show examples of these two kinds of relations in Figure~\ref{fig:running-example}. When two entities have mentions co-occurred in the same sentence, they may express intra-sentential relations. Otherwise, they may express inter-sentential relations. 

% 动机一：句内关系和句间关系需要分开来考虑。
% 常老师的意见：需要从语言学的角度拔高论点，分别说明intra和inter所需要的信息具体是什么, 并且需要为模型部分做铺垫。
Previous methods do not explicitly distinguish these two kinds of relations in the design of the model and use the same method to represent them. 
However, from the perspective of linguistics, intra-sentential relations and inter-sentential relations are expressed in different patterns. For two intra-sentential entities, their relations are usually expressed from local patterns within their co-occurred sentences. As shown in the first example in Figure~\ref{fig:running-example}, \textit{(Polar Music, country of origin, Swedish)} and \textit{(Wembley Arena, located in, London)} can be inferred based solely on the sentence they reside in, i.e., sentences $1$ and $6$ respectively. Unlike intra-sentential relations, inter-sentential relations tend to be expressed from the global interactions across multiple related sentences, also called supporting evidence. Moreover, cross-sentence relations usually require complex reasoning skills, e.g., logical reasoning. As shown in the second example in Figure~\ref{fig:running-example}, (\textit{São Paulo}, continent, \textit{South America}) can be inferred from the other two relation facts expressed in the document: (\textit{São Paulo}, country, \textit{Brazil}) and (\textit{Brazil}, continent, \textit{South America}). So the different patterns between intra- and inter-sentential relations show that it would be better for a model to treat intra- and inter-sentential relations differently. However, previous works usually use the information from the whole document to represent all relations, e.g., 13 sentences for predicting \textit{(Polar Music, country of origin, Swedish)} in the first example in Figure~\ref{fig:running-example}. We argue that this will bring useless noises from unrelated sentences that misguide the learning of relational patterns. 
% To find if this is non-trivial in the existing dataset, we analyze the largest doc-level RE dataset DocRED~\citep{DocRED-paper} and find in dev set that 93.4$\%$ of intra-sentential relations can be extracted based only on their co-occurred sentence, \footnote{DocRED contains the supporting evidence annotated for each relation instance, i.e., sentences that annotators think is useful for inferring this relation instance, as shown in Figure~\ref{fig:running-example}.} which is as expected.

% logical reasoning among relation instances is a very thorny problem in doc-level RE. As shown in the second example in Figure~\ref{fig:running-example}, the inter-sentential relation (\textit{São Paulo}, continent, \textit{South America}) can be inferred from the other two relation facts expressed in the document: (\textit{São Paulo}, country, \textit{Brazil}) and (\textit{Brazil}, continent, \textit{South America}). 

% 动机二：前人的逻辑推理依旧是隐式的，而且不能cover所有情况。
Besides, previous methods~\citep{EoG,GAIN} treat logical reasoning as a representation learning problem. They construct a document graph from the input document using entities as nodes. And the paths between two entities on their graphs, usually passing through other entities, could be regarded as clues for logical reasoning. 
% But since all relations are predicted simultaneously, they are still an implicit way of logical reasoning without the nature of the process. So we should consider an explicit way of logical reasoning where predicting a relation is based on already predicted other relational facts or relational representations.
However, since not all entity pairs can be connected with a path and have the correct logical reasoning paths available on the graph, many cases of logical reasoning cannot be covered. So their methods are somehow limited, and we should consider a new form of logical reasoning to better model and cover all possible reasoning chains.

% 为了解决以上两点前人的不足，我们提出模型来针对性地解决。
In this paper, we propose a novel architecture called \textbf{S}eparate \textbf{I}ntra- and inter-sentential \textbf{RE}asoning (\textbf{SIRE}) for doc-level RE. Unlike previous works in this task, we introduce two different methods to represent intra- and inter-sentential relations respectively. 
For an intra-sentential relation, we utilize a sentence-level encoder to represent it in every co-occurred sentence. Then we get the final representation by aggregating the relational representations from all co-occurred sentences. This will encourage intra-sentential entity pairs to focus on the local patterns in their co-occurred sentences. 
For an inter-sentential relation, we utilize a document-level encoder and a mention-level graph proposed by~\citet{GAIN} to capture the document information and interactions among entity mentions, document, and local context. Then, we apply an evidence selector to encourage inter-sentential entity pairs to selectively focus on the sentences that may signal their cross-sentence relations, i.e., finding supporting evidence.
Finally, we develop a new form of logical reasoning module where one relation instance can be modeled by attentively fusing the representations of other relation instances in all possible logical chains. This form of logical reasoning could cover all possible cases of logical reasoning in the document.

% 贡献。
Our contributions can be summarized as follows:
\begin{itemize}
    \item We propose an effective architecture called SIRE that utilizes two different methods to represent intra-sentential and inter-sentential relations for doc-level RE.
    \item We come up with a new and straightforward form of logical reasoning module to cover all cases of logical reasoning chains.
\end{itemize}

% 实验结论。
We evaluate our SIRE on three public doc-level RE datasets. Experiments show SIRE outperforms the previous state-of-the-art models. Further analysis shows SIRE could produce more reliable and explainable predictions which further proves the significance of the separate encoding.

%% file: files/model.tex
\section{Separate Intra- and Inter-sentential Reasoning (SIRE) Model}

\begin{figure*}
    \centering
    \includegraphics[width=1.0\linewidth]{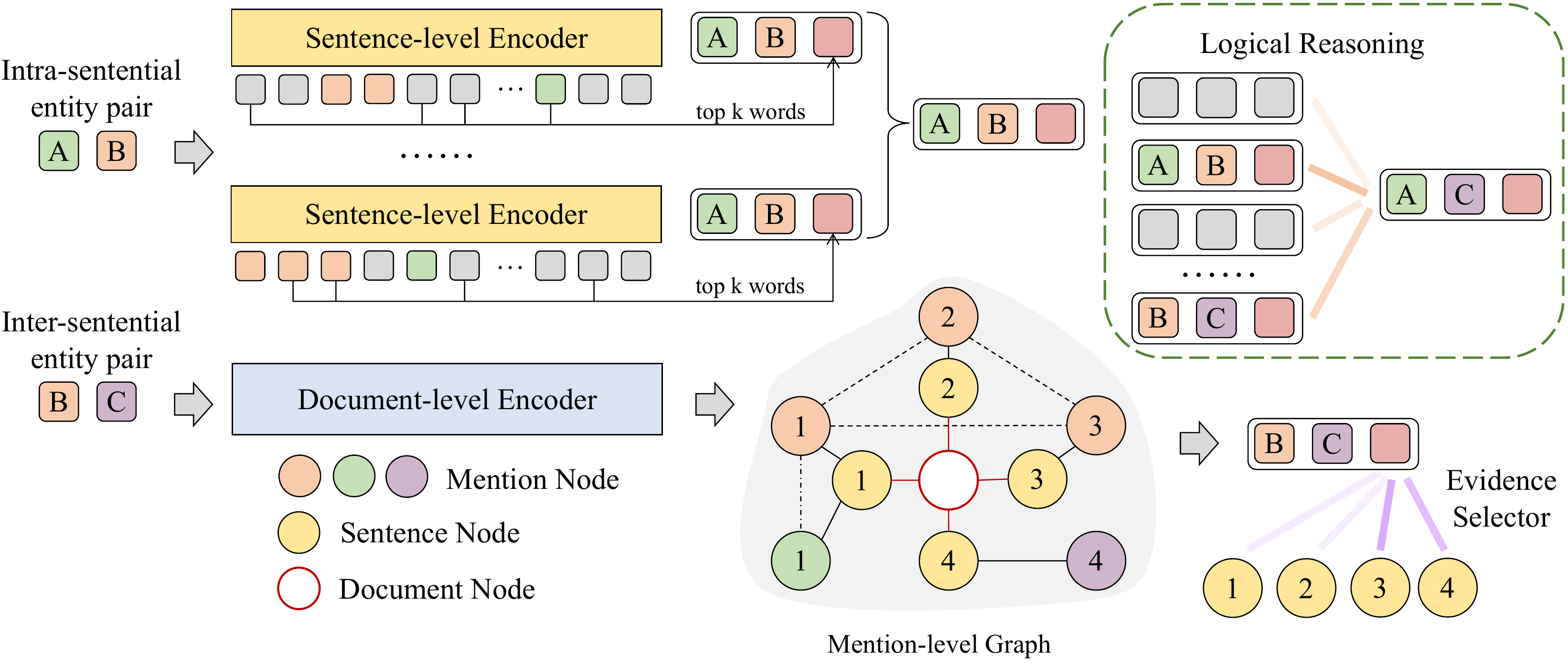}
    \caption{The architecture of SIRE. In the mention-level graph, the number in each circle is its sentence number. Mention nodes with the same color belong to the same entity. Different types of edges are in different styles of line. Our model uses different methods to represent intra- and inter-sentential relations and the self-attention mechanism to model the logical reasoning process. We use the logical reasoning chain:$e_A\rightarrow e_B\rightarrow e_C$ for illustration.}
    \label{fig:model}
\end{figure*}

% SIRE mainly consists of five modules: encoding module (Sec.~\ref{ssec:encoding}), augmented mention-level graph module (Sec.~\ref{ssec:MG}), intra- and inter-sentential relation representation module (Sec.~\ref{ssec:intra-inter}), logical reasoning module (Sec.~\ref{ssec:reasoning}), classification module (Sec.~\ref{ssec:classification}), as is shown in Figure~\ref{fig:model}.
SIRE mainly consists of three modules: intra- and inter-sentential relation representation module (Sec.~\ref{ssec:intra-inter}), logical reasoning module (Sec.~\ref{ssec:reasoning}), classification module (Sec.~\ref{ssec:classification}), as is shown in Figure~\ref{fig:model}. Assume we have a document $\mathcal{D}$ containing $l$ sentences $\left\{\mathcal{S}_{i}\right\}^{l}_{i=1}$. 

\subsection{Intra- and Inter-sentential Relation Representation Module \label{ssec:intra-inter}}
As is discussed in Sec.~\ref{sec:introduction}, for two intra-sentential entities, their relations are usually determined by the local patterns from their co-occurred sentences, while for two inter-sentential entities, their relations are usually expressed across multiple related sentences that can be regarded as the supporting evidence for their relations. So in this module, we utilize two different methods to represent intra-sentential and inter-sentential relations separately. Our methods encourage intra-sentential entity pairs to focus on their co-occurred sentences as much as possible and encourage inter-sentential entity pairs to selectively focus on the sentences that may express their cross-sentence relations. We use three parts to represent the relation between two entities: head entity representation, tail entity representation and context representation.

\subsubsection{Intra-sentential Relation Representation Module\label{sssec:intra}}
% As we mentioned in Sec.~\ref{sec:introduction}, intra-sentential relations could be inferred with a high probability based solely on the sentences in which they co-occur. So we use the entity mentions in their co-occurred sentences to represent the head and tail entity and use the related words in their co-occurred sentences as the context representation.

% intra的编码 (sentence-level encoder)
\textbf{Encoding.} We use a sentence-level encoder to capture the context information for intra-sentential relations and produce contextualized word embedding for each word. Formally, we convert the $i$-th sentence $\mathcal{S}_{i}$ containing $n_i$ words $\left\{w^{\mathcal{S}_{i}}_j\right\}^{n_i}_{j=1}$ into a sequence of vectors $\left\{\textbf{g}^{\mathcal{S}_{i}}_{j}\right\}^{n_i}_{j=1}$.

For each word $w$ in $\mathcal{S}_{i}$, we first concatenate its word embedding with entity type embedding and co-reference embedding\footnote{The existing doc-level RE datasets annotate which mentions belong to the same entity. So for each word in the document, it may belong to the $i$-th entity or non-entity in the document. We embed this co-reference information between entity mention (surface words) and entity (an abstract concept) into the initialized representation of a word.}:
\begin{equation}
\textbf{x} = [E_w(w) ; E_t(t); E_c(c)]
\label{embedding}
\end{equation}
where $E_w(\cdot)$ , $E_t(\cdot)$ and $E_c(\cdot)$ denote the word embedding layer, entity type embedding layer and co-reference embedding layer, respectively. $t$ and $c$ are named entity type and entity id.\footnote{For those words not belonging to any entity, we introduce \textit{None} entity type and id.}

Then the vectorized word representations are fed into the sentence-level encoder to obtain the sentence-level context-sensitive representation for each word:
\begin{equation}
[\textbf{g}^{\mathcal{S}_{i}}_{1}, \ldots, \textbf{g}^{\mathcal{S}_{i}}_{n_{i}}] = f_{enc}^{\mathcal{S}}([\textbf{x}^{\mathcal{S}_{i}}_1, \ldots, \textbf{x}^{\mathcal{S}_{i}}_{n_{i}}])
\end{equation}
where the ${f}_{enc}^{\mathcal{S}}$ denotes sentence-level encoder, which can be any sequential encoder. We will also get the sentence representation $\textbf{s}^{\mathcal{S}_i}$ for sentence $\mathcal{S}_i$ from this encoder. For LSTM, $\textbf{s}^{\mathcal{S}_i}$ is the hidden state of the last time step; for BERT, $\textbf{s}^{\mathcal{S}_i}$ is the output representation of the special marker [CLS]. \\
% intra的表示
\noindent
\textbf{Representing.} For $i$-th entity pair $(e_{i,h}, e_{i,t})$ which expresses intra-sentential relations, where $e_{i,h}$ is the head entity and $e_{i,t}$ is the tail entity, their mentions co-occur in $C$ sentences $S_{co-occur} = \left\{\mathcal{S}_{i_1}, \mathcal{S}_{i_2}, \ldots,  \mathcal{S}_{i_C}\right\}$ once or many times. 
In $j$-th co-occurred sentence $\mathcal{S}_{i_j}$, we use the entity mentions in $\mathcal{S}_{i_j}$ to represent head and tail entity. And we define that the context representation of this relation instance in $\mathcal{S}_{i_j}$ is the top $K$ words correlated with the relations of these two mentions.

Specifically, head entity mention ranging from $s$-th to $t$-th word is represented as the average of the words it contains: $\textbf{e}^{\mathcal{S}_{i_j}}_{i,h} = \frac{1}{t-s+1} \sum^{t}_{k=s}\textbf{g}^{\mathcal{S}_{i_j}}_{k}$, so is the tail entity mention $\textbf{e}^{\mathcal{S}_{i_j}}_{i,t}$. 
Then, we concatenate the representations of head and tail entity mentions and use it as a query to attend all words in $\mathcal{S}_{i_j}$ and compute relatedness score for each word in $\mathcal{S}_{i_j}$ :
\begin{equation}
s_{i,k} = \sigma((W_{intra} \cdot [\textbf{e}^{\mathcal{S}_{i_j}}_{i,h}; \textbf{e}^{\mathcal{S}_{i_j}}_{i,t}])^T \cdot \textbf{g}^{\mathcal{S}_{i_j}}_k)
\end{equation}
% \alpha_{i,k} = \frac{e^{s_{i,k}}}{\sum_{l}e^{s_{i,l}}} 
\begin{equation}
    \alpha_{i,k} = Softmax(s_{i,k})
\end{equation}
where $[\cdot;\cdot]$ is a concatenation operation. $W_{intra} \in \mathbb{R}^{d \times 2d}$ is a parameter matrix. $\sigma$ is an activation function (e.g., ReLU).

Then, we average the representations of top $K$ related words to represents the context information $\textbf{c}_{i}$ for intra-sentential entity pair $(e_{i,h}, e_{i,t})$ in $\mathcal{S}_{i_j}$. In order to make $W_{intra}$ trainable during computing gradient, we also add an item which is the weighted average representation of all words:
\begin{equation}
    \textbf{c}^{\mathcal{S}_{i_j}}_{i} = \beta \cdot \frac{1}{K}\sum_{k \in topK(\alpha_{i,*})} \textbf{g}^{\mathcal{S}_{i_j}}_k + (1 - \beta) \cdot \sum^{n_{i_j}}_{t} \alpha_{i,t} \textbf{g}^{\mathcal{S}_{i_j}}_t
    \label{equation:topk}
\end{equation}
where $\beta$ is a hyperparameter and we use $0.9$ here to force model to focus on the top $K$ words but still consider the subtle influence from other words.

Next, we concatenate the three parts obtained above to form the relational representation of intra-sentential entity pair $(e_{i,h}, e_{i,t})$ in $\mathcal{S}_{i_j}$ and further average the representations in all co-occured sentences $\mathcal{S}_{co-occur}$ to get our final relation representation $\textbf{r}_{i}$ for intra-sentential entity pair $(e_{i,h}, e_{i,t})$ \footnote{If a head entity mentioned $N$ times in a sentence, we will get $N$ intra-sentential relational representations for each of the other tail entities in this sentence.}:
\begin{equation}
    \textbf{r}_{i} = \frac{1}{C} \sum_{\mathcal{S}_{i_j} \in {\mathcal{S}_{co-occur}}} [\textbf{e}^{\mathcal{S}_{i_j}}_{i,h}; \textbf{e}^{\mathcal{S}_{i_j}}_{i,t}; \textbf{c}^{\mathcal{S}_{i_j}}_{i}]
    \label{eq-6}
\end{equation}

This way, we could force the intra-sentential entity pairs to focus on the semantic information from their co-occurred sentences and ignore the noise information from other sentences.

% Finally, since not all intra-sentential relations can be inferred based only on the sentences they reside in, we enhance the local representation with the global representation, which will cover all cases, to get our final relation representation for intra-sentential entity pair $(e_{i,h}, e_{i,t})$:
% \begin{equation}
%     \textbf{r}_{i} = \beta_{2} \cdot \textbf{r}^{local}_{i} + (1 - \beta_{2}) \cdot \textbf{r}^{global}_{i}
% \end{equation}
% where $[\cdot;\cdot]$ is a concatenation operation. $\beta_{2}$ is a hyperparameter and we choose 0.9 which is the best practice and is inline with the portion we previously analyze.

\subsubsection{Inter-sentential Relation Representation Module\label{sssec:inter}}

% inter的表示（document-level encode + Mention-level Graph）
\textbf{Encoding.} According to the nature of inter-sentetential relation, we use a document-level encoder to capture the global interactions for inter-sentential relations and produce contextualized word embedding for each word. Formally, we convert a document $\mathcal{D}$ containing $m$ words  $\left\{w^{\mathcal{D}}_i\right\}^{m}_{i=1}$ into a sequence of vectors $\left\{\textbf{{g}}^{\mathcal{D}}_{j}\right\}^{m}_{j=1}$.

Same as the embedding for intra-sentential relations, we use Equation~\ref{embedding} to embed each word in the document. Then the vectorized word representations are fed into the document-level encoder to obtain document-level context-sensitive representation for each word:
\begin{equation}
[\textbf{g}^{\mathcal{D}}_{1}, \ldots, \textbf{g}^{\mathcal{D}}_{m}] = f_{enc}^{\mathcal{D}}([\textbf{x}^{\mathcal{D}}_1, \ldots, \textbf{x}^{\mathcal{D}}_{m})
\end{equation}
where ${f}_{enc}^{\mathcal{D}}$ denotes the document-level encoder. And we will also get the document representation $\textbf{d}^{\mathcal{D}}$ from this encoder.

To further enhance the document interactions, we utilize the mention-level graph (MG) proposed by \citet{GAIN}. MG in ~\citet{GAIN} contains two different nodes: mention node and document node. Each mention node denotes one particular mention of an entity. Furthermore, MG also has one document node that aims to model the document information. We argue that this graph only contains nodes concerning prediction, i.e., the mentions of the entities and document information. However, it does not contain the local context information, which is crucial for the interaction among entity mentions and the document. So we introduce a new type of node: sentence node and its corresponding new edges to infuse the local context information into MG.

So there are four types of edges\footnote{Note that we remove the mention-document edges of original MG in~\citep{GAIN} and substitute them by introducing mention-sentence and sentence-document edges.} in MG: \\
\noindent
\textbf{Intra-Entity Edge:} Mentions referring to the same entity are fully connected. This models the interactions among mentions of the same entity.\\
\noindent
\textbf{Inter-Entity Edge:} Mentions co-occurring in the same sentence are fully connected. This models the interactions among different entities via co-occurrences of their mentions. \\
\noindent
\textbf{Sentence-Mention Edge:} Each sentence node connects with all entity mentions it contains. This models the interactions between mentions and their local context information. \\
% \noindent
% \textbf{Sentence-Sentence Edge:} All sentence nodes are fully connected. \\
% \noindent
% \textbf{Mention-Document Edge:} All mention nodes are connected to the document node. \\
\noindent
\textbf{Sentence-Document Edge:} All sentence nodes are connected to the document node. This models the interactions between local context information and document information, acting as a bridge between mentions and document.

Next, we apply Relational Graph Convolutional Network (R-GCN, \citealp{RGCN}) on MG to aggregate the features from neighbors for each node. Given node $u$ at the $l$-th layer, the graph convolutional operation can be defined as:
\begin{equation}
       \textbf{h}_{u}^{(l + 1)} = ReLU \left(\sum_{t\in\mathcal{T}}\sum_{v\in\mathcal{N}^{t}_{u}\bigcup \{u\}}  \frac{1}{c_{u,t}}W^{(l)}_t \textbf{h}_{v}^{(l)}\right)
\end{equation}
where $\mathcal{T}$ is a set of different types of edges, $W^{(l)}_t\in \mathbb{R}^{d\times d}$ is a trainable parameter matrix.
$\mathcal{N}^{t}_{u}$ denotes a set of neighbors for node $u$ connected with $t$-th type edge. $c_{u,t} = |\mathcal{N}^{t}_{u}|$ is a normalization constant.

We then aggregate the outputs of all R-GCN layers to form the final representation of node $u$:
\begin{equation}
    \textbf{m}_u = ReLU(W_u \cdot [\textbf{h}_{u}^{(0)}; \textbf{h}_{u}^{(1)}; \ldots; \textbf{h}_{u}^{(N)}])
\end{equation}
where $W_u \in \mathbb{R}^{d\times Nd}$ is a trainable parameter matrix. $h^{(0)}_u$ is the initial representation of node $u$. For a mention ranging from the $s$-th word to the $t$-th word in the document, $\textbf{h}^{(0)}_u = \frac{1}{t-s+1} \sum_{j=s}^{t} \textbf{g}^{\mathcal{D}}_{j}$; for $i$-th sentence node, it is initialized with $\textbf{s}^{\mathcal{S}_i}$ from sentence-level encoder; for the document node, it is initialized with $\textbf{d}^{\mathcal{D}}$ from document-level encoder.

% Now, we can get the global representation for every relation instance, including intra-sentential and inter-sentential relations. \citet{THU_RE} find that there are two primary information sources for predicting relations between entities: textual context and entity mentions themselves. So we here use three parts to represent a relation instance at the global level: head entity representation, tail entity representation, context representation.

% For the $i$-th relation instance, head and tail entity representation are defined as the average of their entity mentions from MG:
% \begin{equation}
%     \textbf{e}^{global}_{i} = \frac{1}{N} \sum_{j \in M(e_i)} \textbf{m}_j
%     \label{equation:global-entity}
% \end{equation}
% where the $M(e_i)$ is the mention set of $e_i$. And the context representation is the document node from MG: $\textbf{c}^{global}_{i} = \textbf{m}_{global}$. So the final global representation of $i$-th relation instance is:
% \begin{equation}
%     \textbf{r}^{global}_{i} = [\textbf{e}^{global}_{i,h};\textbf{e}^{global}_{i,t};\textbf{c}^{global}_{i}]
% \end{equation}
% where $[\cdot;\cdot]$ is a concatenation operation.
\noindent
\textbf{Representing.} We argue that inter-sentential relations can be inferred from the following information sources: 1) the head and tail entities themselves; 2) the related sentences that signal their cross-sentence relations, namely supporting evidences; 3) reasoning information such as logical reasoning, co-reference reasoning, world knowledge, etc. We here only consider the first two information and leave the last in Sec.~\ref{ssec:reasoning}.

Different from intra-sentential relations, inter-sentential relations tend to be expressed from the global interactions. So for the $i$-th entity pair $(e_{i,h}, e_{i,t})$ which expresses inter-sentential relation, the head entity representation $\textbf{e}_{i,h}$ and the tail entity representation and $\textbf{e}_{i,t}$ are defined as the average of their entity mentions from MG:
\begin{equation}
    \textbf{e}_{i} = \frac{1}{N} \sum_{j \in M(e_i)} \textbf{m}_j
    \label{equation:global-entity}
\end{equation}
where the $M(e_i)$ is the mention set of $e_i$.

And we apply an evidence selector with attention mechanism \citep{Attention} to encourage the inter-sentential entity pair to selectively focus on the sentences that express their cross-sentence relations. This process could be regarded as finding supporting evidence for their relations. So the context representation $\textbf{c}_{i}$ for inter-sentential entity pair $(e_{i,h}, e_{i,t})$ is the weighted average of the sentence representations from MG:
\begin{equation}
P({\mathcal{S}_{k}|e_{i,h},e_{i,t})} = \sigma (W_k \cdot [\textbf{e}_{i,h}; \textbf{e}_{i,t}; \textbf{m}_{\mathcal{S}_{k}}])
\label{equation:inter-context-logits}
\end{equation}
\begin{equation}
    \alpha_{i,k} = \frac{P(\mathcal{S}_{k}|e_{i,h},e_{i,t})}{\sum_{l}P(\mathcal{S}_{l}|e_{i,h},e_{i,t})}
    \label{equation:inter-context-percent}
\end{equation}
\begin{equation}
    \textbf{c}_{i} = \sum^{l}_{k}\alpha_{i,k} \cdot \textbf{m}_{\mathcal{S}_{k}}
\end{equation}
where $W_k\in \mathbb{R}^{1 \times 2d}$ is a trainable parameter matrix. $\sigma$ is a $sigmoid$ function. 
% Similarly, the context information for inter-sentential relations may also be expressed from the whole document information, so we add an item to encourge the inter-sentential relations to focus on both sentence information and document information.
% \begin{equation}
% \textbf{c}_{i} = 
% \beta_{3} \cdot \textbf{c}^{local}_{i} + (1 - \beta_{3}) \cdot \textbf{m}_{doc}
% \end{equation}
% where $\beta_{3}$ is a hyperparameter.

Next, the final relation representation for inter-sentential entity pair $(e_{i,h}, e_{i,t})$ should be:
\begin{equation}
\textbf{r}_{i} = [\textbf{e}_{i,h}; \textbf{e}_{i,t}; \textbf{c}_{i}]
\label{eq-14}
\end{equation}

% Equation~\ref{equation:inter-context-logits} and ~\ref{equation:inter-context-percent} can be regarded as to find supporting evidence for inter-sentential relation instance. So we naturally introduce the evidence BCE loss, where the labels of supporting evidence are provided by DocRED annotation, to facilitate the learning of sentence selection for inter-sentential relations:
% \begin{equation}
%     \begin{aligned}
%     \mathcal{L}_{evi} = -\sum_{\mathcal{S}_k\in \mathcal{D}} y_s \cdot \log P(\mathcal{S}_{k}|e_{i,h},e_{i,t}) + & \\ (1 - y_s) \cdot \log (1 - P(\mathcal{S}_{k}|e_{i,h},e_{i,t}))
%     \end{aligned}
% \end{equation}

\subsection{Logical Reasoning Module\label{ssec:reasoning}}
In this module, we focus on logical reasoning modeling. As mentioned in Sec.~\ref{sec:introduction}, previous works usually use the paths between each entity pair as the clues for logical reasoning. Furthermore, they concatenate the path representations with entity pair representations to predict relations. However, since not all entity pairs are connected with a path and have the correct logical reasoning paths in their graph, many cases of logical reasoning cannot be covered. So their methods are somehow limited.

In this paper, we utilize self-attention mechanism \citep{transformer} to model logical reasoning. Specifically, we can get the relational representations for all entity pairs from the above sections. For $i$-th entity pair $(e_{h}, e_{t})$, we can assume there is a two-hop logical reasoning chains: $e_{h} \rightarrow e_{k} \rightarrow e_{t}$ in the document, where $e_k$ can be any other entities in the document except $e_h$ and $e_t$. So $(e_{h}, e_{t})$ can attend to all the relational representations of other entity pairs including $(e_{h}, e_{k})$ and $(e_{k}, e_{t})$, termed as $\mathcal{R}_{att}$. Finally, the weighted sum of $\mathcal{R}_{att}$ can be treated as a new relational representation for $(e_{h}, e_{t})$, which considers all possible two-hop logical reasoning chains in the document.\footnote{This can be scaled to muti-hop logical reasoning by increasing the self-attention layers. We only consider two-hop logical reasoning in this paper following ~\citet{GAIN}.}
\begin{equation}
    \textbf{r}^{new}_{i} = \sum_{\textbf{r}_k \in \mathcal{R}_{att} \cup \{\textbf{r}_i\}} \gamma_{k} \cdot \textbf{r}_{k}
\end{equation}
\begin{equation}
    \gamma_{k} = Softmax((W_{att} \cdot \textbf{r}_{i})^T \cdot \textbf{r}_k)
\end{equation}
where  $W_{att} \in \mathbb{R}^{3d \times 3d}$ is a parameter matrix.

In this way, the path in the previous works could be converted into the individual attention on every entity pair in the logical reasoning chains. We argue that this form of logical reasoning is simpler and more scalable because it will consider all possible logical reasoning chains without connectivity constraints in the graph structure.

\subsection{Classification Module\label{ssec:classification}}
We formulate the doc-level RE task as a multi-label classification task:
\begin{equation}
    P(r|e_{i,h}, e_{i,t}) = sigmoid \left(W_1 \sigma (W_2 \textbf{r}_{i} + b_1) + b_2 \right)
\end{equation}
where $W_1$, $W_2$, $b_1$, $b_2$ are trainable parameters, $\sigma$ is an activation function (e.g., ReLU). We use binary cross entropy as objective to train our SIRE:
\begin{equation}
\begin{split}
    \mathcal{L}_{rel} &= - \sum _{\mathcal{D} \in \mathcal{C}}  \sum _{h\neq t} \sum_{r_i \in \mathcal{R}}  \mathbb I \left ( r_i=1 \right ) \log P \left ( r_i|e_{i,h}, e_{i,t} \right )
    \\
     & +  \mathbb I \left ( r_i=0 \right ) \log \left(  1 - P \left ( r_i|e_{i,h}, e_{i,t} \right ) \right )
\end{split}
\end{equation}
where $\mathcal{C}$ denotes the whole corpus, $\mathcal{R}$ denotes relation type set and $\mathbb I\left(\cdot\right)$ refers to indicator function.

%% file: files/experiments.tex
\section{Experiments}

\begin{table*}[htbp]
\centering
% \scriptsize
\small
\begin{tabular}{lcccccc}
\hline
\multirow{2}{*}{\bf Model} & \multicolumn{4}{c}{\bf Dev} & \multicolumn{2}{c}{\bf Test}  \\ 
\cmidrule(lr){2-5} \cmidrule(lr){6-7}
~ & \bf Ign F1 & \bf F1 & \bf Intra-F1 & \bf Inter-F1 & \bf Ign F1 & \bf F1 \\
\hline

BiLSTM \citep{DocRED-paper} & 48.87 & 50.94 & 57.05 & 43.49 & 48.78 & 51.06 \\
HIN-GloVe \citep{HIN} & 51.06 & 52.95 & - & - & 51.15 & 53.30 \\
LSR-GloVe \citep{LSR} & 48.82 & 55.17 & 60.83 & 48.35 & 52.15 & 54.18 \\
GAIN-GloVe \citep{GAIN} & 53.05 & 55.29 & 61.67 & 48.77 & 52.66 & 55.08 \\
\hline
SIRE-GloVe & \textbf{54.10} & \textbf{55.91} & \textbf{62.94} & \textbf{48.97} & \textbf{54.04} & \textbf{55.96} \\
\quad \textit{-LR Module} & 53.73 & 55.58 & 62.77 & 47.87 & 53.75 & 55.55 \\
\quad \textit{-context} & 52.57 & 54.41 & 61.66 & 46.92 & 52.33 & 54.15 \\
\quad \textit{-inter4intra} & 52.23 & 54.26 & 60.81 & 48.36 & 51.77 & 53.30 \\
\hline 
\hline
BERT \citep{finetune-bert} & - & 54.16 & 61.61 & 47.15 & - & 53.20 \\
BERT-Two-Step \citep{finetune-bert} & - & 54.42 & 61.80 & 47.28 & - & 53.92 \\
HIN-BERT \citep{HIN} & 54.29 & 56.31 & - & - & 53.70 & 55.60 \\
CorefBERT \citep{CorefBERT} & 55.32 & 57.51 & - & - & 54.54 & 56.96 \\
GLRE-BERT \citep{GLRE} & - & - & - & - & 55.40 & 57.40 \\
LSR-BERT \citep{LSR} & 52.43 & 59.00 & 65.26 & 52.05 & 56.97 & 59.05 \\
GAIN-BERT \citep{GAIN} & 59.14 & 61.22 & 67.10 & 53.90 & 59.00 & 61.24 \\
\hline
SIRE-BERT & \textbf{59.82} & \textbf{61.60} & \textbf{68.07} & \textbf{54.01} & \textbf{60.18} & \textbf{62.05} \\
% \quad \textit{-LR Module} & 59.50 & 61.42 & 67.89 & 52.44 & 59.88 & 61.86 \\
\hline
\end{tabular}
\caption{Performance on DocRED. Models above the double line do not use pre-trained model. \textit{LR Module} is the logical reasoning module. \textit{context} denotes context representations in Eq.~\ref{eq-6} and Eq.~\ref{eq-14}. \textit{inter4intra} denotes using the inter-sentential module also for intra-sentential entity pairs.
% \footnotemark[1]
}
\label{table:results}
\end{table*}

\begin{table}[htbp]
\centering
\small
\begin{tabular}{lcccccc}
\hline
\bf Model  & \bf CDR  & \bf GDA \\ 
\hline
BRAN \citep{BRAN} & 62.1 & - \\
EoG \citep{GLRE} & 63.6 & 81.5 \\
LSR \citep{LSR} & 64.8 & 82.2 \\
GLRE-BioBERT \citep{GLRE} & 68.5 & - \\
\hline
SIRE-BioBERT & \textbf{70.8} & \textbf{84.7} \\
\hline
\end{tabular}
\caption{Performance on CDR and GDA.
% \footnotemark[1]
}
\label{table:results2}
\end{table}

% \begin{table*}
% \centering
% \begin{tabular}{lll}
% \hline \textbf{Datasets} & \textbf{# rel. type} & \textbf{# ent. type} & \textbf{# docs.} & \textbf{#avg. sent.} & \textbf{}\\ 
% \hline
% DocRED & 96 & 7
% CDR & 1 & 2
% GDA & 1 & 2
% \hline
% \end{tabular}
% \caption{\label{datasets-statistics} Font guide. }
% \end{table*}
\subsection{Dataset}
We evaluate our proposed model on three document-level RE datasets: \\
\textbf{DocRED}: 
The largest human-annotated document-level relation extraction dataset was proposed by ~\citet{DocRED-paper}. It is constructed from Wikipedia and Wikidata and contains $96$ types of relations, $132,275$ entities, and $56,354$ relational facts in total. Documents in DocRED have about $8$ sentences on average. More than $40.7\%$ relation facts can only be extracted from multiple sentences.
$61.1\%$ relation instances require various reasoning skills such as logical reasoning. $93.4\%$ intra-sentential relations can be inferred based solely on their co-occurred sentences. We show two examples from DocRED in Figure~\ref{fig:running-example}. We follow the standard split of the dataset, $3,053$ documents for training, $1,000$ for development, and $1,000$ for testing. 

\noindent
\textbf{CDR} (BioCreative V): The Chemical-Disease Reactions dataset was created by ~\citet{CDR} manually. It contains one type of relation: \textit{Chemical-Induced-Disease} between chemical and disease entities. We follow the standard split of the dataset, $500$ documents for training, $500$ for development, and $500$ for testing. 

\noindent
\textbf{GDA} (DisGeNet): The Gene-Disease-Associations dataset was introduced by ~\citet{GDA}.
It contains one type of relation: \textit{Gene-Induced-Disease} between gene and disease entities.
We use standard split of the dataset, $23,353$ documents for training, $5,839$ for development, and $1,000$ for testing. 

% \noindent
% Table~\ref{datasets-statistics} shows the detailed statistics of these three datasets.

% We evaluate the proposed SIRE on the largest human-annotated document-level relation extraction dataset, DocRED \citep{DocRED-paper}. It is constructed from Wikipedia and Wikidata and contains $96$ types of relations, $132,275$ entities, and $56,354$ relational facts in total. Documents in DocRED have about $8$ sentences on average. More than $40.7\%$ relation facts can only be extracted from multiple sentences.
% $61.1\%$ relation instances require various reasoning skills such as logical reasoning. $93.4\%$ intra-sentential relations can be inferred based solely on their co-occurred sentences. We show two examples from DocRED in Figure~\ref{fig:running-example}. We follow the standard split of the dataset, $3,053$ documents for training, $1,000$ for development, and $1,000$ for testing. 

\subsection{Experimental Settings}
In our SIRE implementation, we use $3$ layers of GCN, use ReLU as our activation function, and set the dropout rate to $0.3$, learning rate to $0.001$.
We train SIRE using AdamW \citep{adamW} as optimizer with weight decay $0.0001$ and implement SIRE under PyTorch \citep{PyTorch} and DGL \citep{DGL} frameworks.

We implement two settings for our SIRE. \textbf{SIRE-GloVe} uses GloVe ($100$d, \citealp{GloVe}) and BiLSTM ($512$d, \citealp{BiLSTM}) as word embedding and encoder, respectively.
\textbf{SIRE-BERT} use BERT-base \citep{BERT} as encoder on DocRED, cased BioBERT-Base v1.1 as the encoder on CDR/GDA, and the learning rate for BERT parameters is set to $1e^{-5}$ and learning rate for other parameters remains $1e^{-3}$.
Detailed hyperparameter settings are in Appendix.

\subsection{Baselines and Evaluation Metrics}

We use the following models as our baselines:

\citet{DocRED-paper} propose the \textbf{BiLSTM} \citep{BiLSTM} as the encoder on DocRED and use the output from the encoder to represent all entity pairs to predict relations. 

\citet{finetune-bert} propose \textbf{BERT} to replace the BiLSTM as the encoder on DocRED. Moreover, they also propose \textbf{BERT-Two-Step}, which first predicts whether two entities have a relation and then predicts the specific target relation. 

\citet{HIN} propose the hierarchical inference networks \textbf{HIN-GloVe} and \textbf{HIN-BERT}, which make full use of multi-granularity inference information including entity level, sentence level, and document level to infer relations.

Similar to \citet{finetune-bert}, \citet{CorefBERT} propose a language representation model called \textbf{CorefBERT} as encoder on DocRED that can capture the coreferential relations in context.

\citet{LSR} propose the \textbf{LSR-GloVe} and \textbf{LSR-BERT} to dynamically induce the latent dependency tree structure to better model the document interactions for prediction. 

% \citet{HeterGSAN-Recon} propose a encoder-classifier-reconstructor model \textbf{HeterGSAN-Recon-GloVe} and \textbf{HeterGSAN-Recon-BERT}to reconstruct the ground-truth path dependencies from the graph representation for doc-level RE

\citet{GLRE} propose a global-to-local network \textbf{GLRE}, which encodes the document information in terms of entity global and local representations as well as context relation representations. 

\citet{GAIN} propose the graph aggregation-and-inference networks \textbf{GAIN-GloVe} and \textbf{GAIN-BERT} which utilize two levels of graph structures: mention-level graph and entity-level graph to capture document interactions and conduct path logical reasoning mechanism, respectively. 

\citet{BRAN} propose a self-attention encoder \textbf{BRAN} to consider interactions across mentions and relations across sentence boundaries.

Following the previous works~\citep{DocRED-paper,GAIN}, we use the F1 and Ign F1 as the evaluation metrics to evaluate the overall performance of a model. The Ign F1 metric calculates F1 excluding the common relation facts in the training and dev/test sets. We also use the intra-F1 and inter-F1 metrics to evaluate a model's performance on intra-sentential relations and inter-sentential relations on the dev set.

\begin{figure*}
    \centering
    \includegraphics[width=0.9\linewidth]{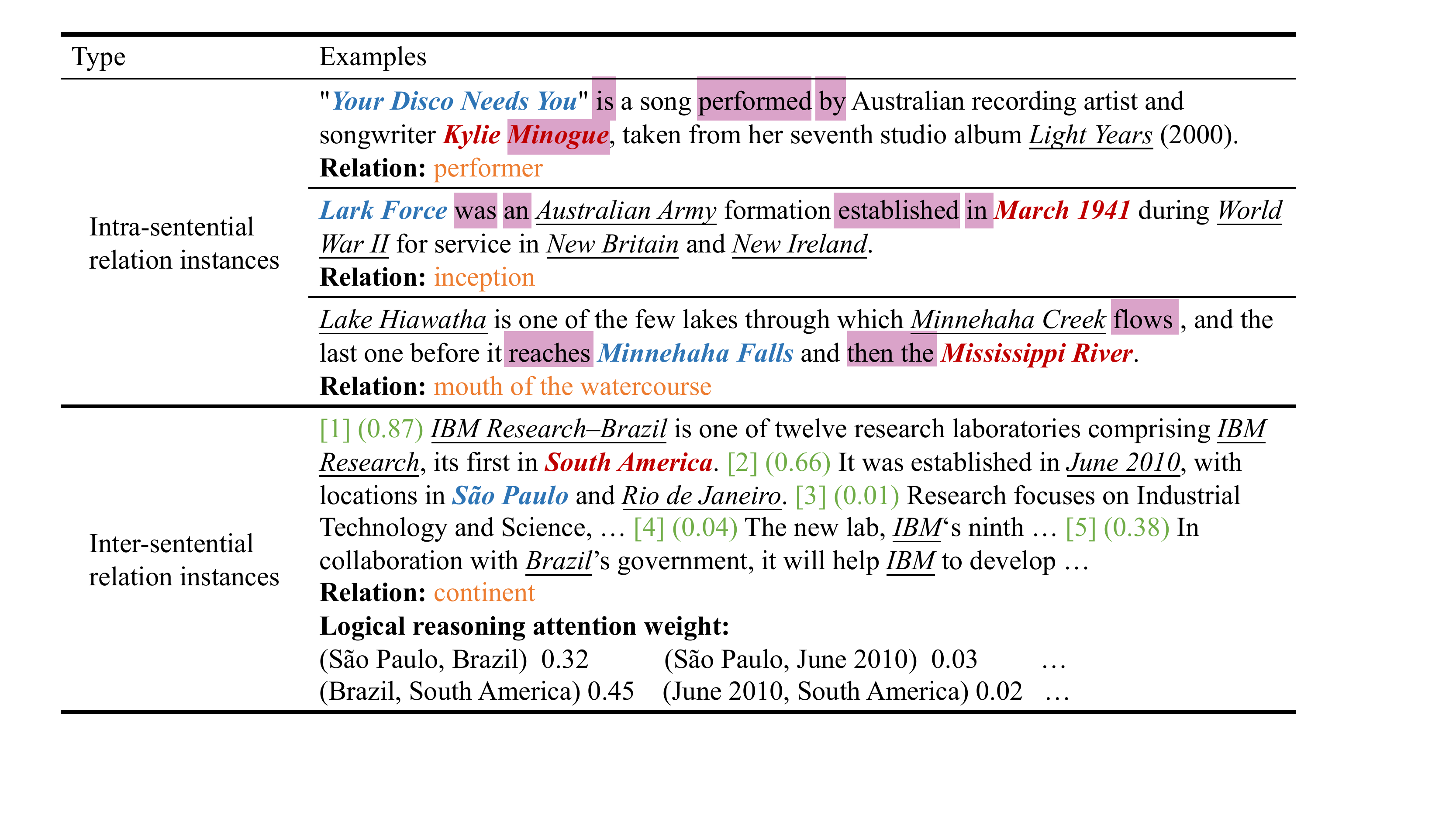}
    \caption{Cases for illustrating the reliable and explainable predictions of our SIRE. Head entities, tail entities, and sentence numbers along with the scores from evidence selector are colored in blue, red, green, respectively. In intra-sentential relations, words with pink background color are the top $4$ words from Equation~\ref{equation:topk}. }
    \label{fig:case}
\end{figure*}

\subsection{Results\label{ssec:quantitative}}
The performances of SIRE and baseline models on the DocRED dataset are shown in Table~\ref{table:results}. Among the model not using BERT encoding, SIRE outperforms the previous state-of-the-art model by 0.88/1.38 F1/Ign F1 on the test set. Among the model using BERT encoding, SIRE outperforms the previous state-of-the-art models by 1.18/0.81 F1/Ign F1 on the test set. The improvement on Ign F1 is larger than that on F1. This shows SIRE has a stronger generalization ability on the unseen relation instances. On intra-F1 and inter-F1, we can observe that SIRE is better than the previous models that indiscriminately represent the intra- and inter-sentential relations in the same way. This demonstrates that representing intra- and inter-sentential relations in different methods is better than representing them in the same way. The improvement on intra-F1 is greater than the improvement on inter-F1. This shows that SIRE mainly improves the performance of intra-sentential relations. 
% This is intuitive due to that previous works use the information from the whole document, e.g., 13 sentences in Figure~\ref{fig:running-example}, to infer intra-sentential relations, which will introduce many noises from other sentences. And our model only considers the information from the co-occurred sentences, which is more reasonable.
The performances of SIRE and baseline models on the CDR/GDA dataset are shown in Table~\ref{table:results2}, which are consistent with the improvement on DocRED.

\subsection{Ablation Study\label{ssec:ablation}}

To further analyze SIRE, we also conduct ablation studies to illustrate the effectiveness of different modules in SIRE. We show the results in Table~\ref{table:results}. \\
\noindent
1) \textbf{the importance of the logical reasoning module}:
When we discard the logical reasoning module, the performance of SIRE-GloVe decreases by 0.41 F1 on the DocRED test set. This shows the effectiveness of our logical reasoning module, which can better model the reasoning information in the document. Moreover, it drops significantly on inter-F1 and drops fewer points on intra-F1. This shows our logical reasoning module mainly improves the performance of the inter-sentential relations that usually require reasoning skills. \\
\noindent
2) \textbf{Ablation on context representations in Eq.~\ref{eq-6} and Eq.~\ref{eq-14}}: When we remove the context representations in intra- and inter-sentential relational representations, the performance of SIRE-GloVe on the DocRED test set drops by 1.81 F1. This shows context information (top K words for intra, evidence sentences for inter) is important for both intra- and inter-sentential relation representation. \\
\noindent
3) \textbf{Using the inter-sentential module also for intra-sentential entity pairs}: In this experiment, we do not distinguish these two types of relations, using the encoding method for inter-sentential to encode all entity pairs, and remain the logical reasoning module unchanged. The performance of SIRE-GloVe drops by 2.66/2.13 F1/intra-F1 on the DocRED test set. This confirms the motivation that we cannot use global information to learn the local patterns for intra-sentential relations.

\subsection{Reasoning Performance \label{ssec:qualitative}}
Furthermore, we evaluate the reasoning ability of our model on the development set in Table~\ref{table:infer}. We use infer-F1 as the metric that considers only two-hop positive relation instances in the dev set. So it will naturally exclude many cases that do not belong to the two-hop logical reasoning process to strengthen the evaluation of reasoning performance. % For example, we take into account the golden relation facts including $e_h {\rightarrow} e_{o} {\rightarrow} e_{t}$ and $e_h {\rightarrow} e_{t}$ when calculating Infer-F1.
As Table~\ref{table:infer} shows, SIRE is superior to previous models in handling the two-hop logical reasoning process. Moreover, after removing the logical reasoning module, out SIRE drops significantly on infer-F1. This shows that our logical reasoning module plays a crucial role in modeling the logical reasoning process.

\begin{table}
\centering
\begin{tabular}{lccc}
\hline
Model & Infer-F1 & P & R \\
\hline
BiLSTM & 38.73 & 31.60 & 50.01 \\
% Context-Aware & 39.73 & 33.97 & 47.85 \\
GAIN-GloVe & 40.82  & 32.76 & 54.14 \\
\hline
SIRE-GloVe & \textbf{42.72}  & \textbf{34.83} & \textbf{55.22}\\
\textit{- LR Module} & 39.18 & 31.97 & 50.59\\
% \hline 
% \hline
% BERT-RE$_{base}$ & 39.62 & 34.12 & 47.23 \\
% RoBERTa-RE$_{base}$ & 41.78 & 37.97 & 46.45 \\
% \hline
% GAIN-BERT$_{base}$ & \textbf{46.89} & \textbf{38.71} & \textbf{59.45}\\
% \textit{- Inference Module} & 45.11 & 36.91 & 57.99\\
\hline
\end{tabular}
\caption{Infer-F1 results on dev set of DocRED. P: Precision, R: Recall.}
\label{table:infer}
\end{table}

\subsection{Case Study\label{ssec:case-study}}

Figure~\ref{fig:case} shows the prediction cases of our SIRE. In intra-sentential relations, the top 4 words related to the relations of three entity pairs conform with our intuition. Our model correctly find the words by using Eq.\ref{equation:topk} that trigger the relations of these entity pairs. In inter-sentential relations, the supporting evidence that the model finds, i.e., sentences $1$ and $2$, indeed expresses the relations between São Paul and South America. We also conduct logical reasoning in terms of the logical reasoning chains: São Paul$\rightarrow$ other-entity $\rightarrow$ South America. Our SIRE could focus on the correct logical reasoning chains: São Paul$\rightarrow$ Brazil $\rightarrow$ South America. These cases show the predictions of SIRE are explainable.
% \subsection{Error Analysis\label{ssec:ablation}}

%% file: files/related_work.tex
\section{Related Work}
\textbf{Document-level relation extraction.} Many recent efforts \citep{quirk-poon-17-distant,peng-etal-17-cross,DBLP:conf/aaai/GuptaRSR19,song-etal-2018,n-ary,DocRED-paper,finetune-bert,HIN,LSR,GAIN,GLRE,CFER} manage to tackle the document-level relation extraction. 
% \citet{HIN} proposed a hierarchical inference network that makes full use of multi-granularity inference information, including entity level, sentence level, and document level. It is a typical work of sequential models on document-level relation extraction.
Most of them use graph-based models, such as Graph Convolutional Networks (GCNs, \citealp{GCN,RGCN}) that has been used in many natural language processing tasks \citep{marcheggiani-titov-2017,DBLP:conf/aaai/YaoM019,liu-etal-2020}. They construct a graph structure from the input document. This graph uses the word, mentions or entities as nodes and uses heuristic rules and semantic dependencies as edges. They use this graph to model document information and interactions and to predict possible relations for all entity pairs. 
\citet{LSR} proposed a latent structure induction to induce the dependency tree in the document dynamically.
\citet{GAIN} proposed a double graph-based graph aggregation-and-inference network that constructs two graphs: mention-level graph and entity-level graph. They use the former to capture the document information and interactions among entity mentions and document and use the latter to conduct path-based logical reasoning.
However, these works do not explicitly distinguish the intra- and inter-sentential relation instances in the design of the model and use the same way to encode them. So the most significant difference between our model and previous models is that we treat intra-sentential and inter-sentential relations differently to conform with the relational patterns for their prediction. \\
\noindent
\textbf{Reasoning in relation extraction.} Reasoning problem has been extensively studied in the field of question answering (\citealp{C-GRU}). However, few works manage to tackle this problem in the document-level relation extraction task.
% \citet{GPNN} proposed a graph neural network with generated parameters where the parameters are produced by a generator taking as inputs natural language sentences.
% Moreover, they proposed a multi-hop Reasoning mechanism by using GCN \citep{GCN} on this graph. 
% We argue that their method could be regarded as implicit relational reasoning due to the generated parameters, while ours is more explicit than theirs.
\citet{GAIN} is the first to propose the explicit way of relational reasoning on doc-level RE, which mainly focuses on logical reasoning. They use the paths on their entity-level graph to provide clues for logical reasoning. However, since not all entity pairs are connected with a path and have the correct logical reasoning paths in their graph, their methods are somehow limited.
In this work, we design a new form of logical reasoning to cover more cases of logical reasoning.
% that can better cope with the logical reasoning challenges.

%% file: files/conclusion.tex
\section{Conclusion}
Intra- and inter-sentential relations are two types of relations in doc-level RE. We propose a novel architecture, SIRE, to represent these two relations in different ways separately in this work. We introduce a new form of logical reasoning module that models logical reasoning as a self-attention among representations of all entity pairs. Experiments show that our SIRE outperforms the previous state-of-the-art methods. The detailed analysis demonstrates that our predictions are explainable. We hope this work will have a positive effect on future research regarding new encoding schema, a more generalizable and explainable model.

%% file: files/appendix.tex
\appendix

\section{Hyperparameter settings\label{sec:appendix}}

We use the development set to manually tune the optimal hyperparameters for SIRE, based on the Ign F1 score.
Experiments are run on NVIDIA-RTX-3090-$24$GB GPU.
% we conducted five runs with different initialized parameters and computed the average performance. 
Hyperparameter settings for SIRE-GloVe, SIRE-BERT on DocRED are listed in Table~\ref{tab:hyperparam1}, ~\ref{tab:hyperparam2}, respectively. The values of hyperparameters we finally adopted are in bold. Note that we do not tune all the hyperparameters.
% \makeatletter\def\@captype{table}\makeatother

% \begin{figure*}
%     \centering
%     \includegraphics[width=1.0\linewidth]{emnlp2020-templates/fig_case_appenddix_Two_Doors_Down.pdf}
%     \caption{Caption}
%     \label{fig:case1}
% \end{figure*}

% \paragraph{Hyperparameter settings:}
% Use {\small\verb|\appendix|} before any appendix section to switch the section numbering over to letters.

\begin{table}[!htbp]
\centering
% \scriptsize
% \small
\begin{tabular}{lr}
\hline
\textbf{Hyperparameter} & Value \\
\hline
Batch Size &  16, \textbf{32} \\
Learning Rate  & \textbf{0.001}  \\
Activation Function & \textbf{ReLU}, Tanh \\
Positive v.s. Negative Ratio & 1, 0.5, \textbf{0.25} \\
Word Embedding Size  & \textbf{200}\\
Entity Type Embedding Size & \textbf{20}  \\
Coreference Embedding Size & \textbf{20}  \\
Encoder Hidden Size &  256, \textbf{512} \\
Dropout &  0.3, \textbf{0.5}, 0.7 \\
Layers of GCN & 1, 2, \textbf{3} \\
GCN Hidden Size & \textbf{1024} \\
% Warm-up steps & - & - & - \\
Weight Decay & \textbf{0.0001} \\
$\beta$ & \textbf{0.9} \\
% Adam $\epsilon$ & - & - & - \\
% Adam $\beta_{1}$ & - & - & - \\
% Adam $\beta_{2}$ & - & - & - \\
% Gradient clipping & - & - & - \\
% Learning rate annealing  &  \\
\hline \hline
Numbers of Parameters & 95M \\ % e.g. 110M
Training Time & 18 hours \\
Hyperparameter Search Trials & 20 \\
\hline
\end{tabular}
\caption{Settings for SIRE-GloVe.}
\label{tab:hyperparam1}
\end{table}

\begin{table}[!htbp]
\centering
% \scriptsize
% \small
\begin{tabular}{lr}
\hline
\textbf{Hyperparameter} & Value \\
\hline
Batch Size & 16, \textbf{32} \\
Learning Rate  & \textbf{0.001}  \\
Activation Function & \textbf{ReLU}, Tanh \\
Positive v.s. Negative Ratio & 1, 0.5, \textbf{0.25} \\
% Word Embedding Size  & \textbf{768}  \\
Entity Type Embedding Size & \textbf{128} \\
Coreference Embedding Size & \textbf{128} \\
% Encoder Hidden Size & \textbf{768} \\
Dropout & \textbf{0.3}, 0.5, 0.7 \\
Layers of GCN & 1, 2, \textbf{3} \\
GCN Hidden Size & \textbf{1024} \\
Weight Decay & \textbf{0.0001} \\
$\beta$ & \textbf{0.9} \\
\hline \hline
Numbers of Parameters & 307M \\ % e.g. 110M
Training Time & 24 hours \\
Hyperparameter Search Trials & 30 \\
\hline
\end{tabular}
\caption{Settings for SIRE-BERT.}
\label{tab:hyperparam2}
\end{table}